\documentclass[onecolumn,table]{wlscirep}
\usepackage[utf8]{inputenc}
\usepackage[T1]{fontenc}

\usepackage{amsmath}
\usepackage{graphicx}
\usepackage[colorlinks=true, allcolors=blue]{hyperref}
\usepackage{tcolorbox}
\usepackage[normalem]{ulem}
\usepackage{xcolor}

\usepackage[english]{babel}

\title{Self-Normalized Density Map (SNDM) for Counting Microbiological Objects}

\author[1,*,+]{Krzysztof M. Graczyk}
\author[2,3,+]{Jarosław Pawłowski}
\author[2,3]{Sylwia Majchrowska}
\author[2]{Tomasz Golan}
\affil[1]{ University of Wroclaw, Institute for Theoretical Physics,\newline
pl. Maxa Borna 9, 50-343 Wroclaw, Poland}

\affil[2]{NeuroSYS, Rybacka 7, 53-656 Wrocław, Poland}

\affil[3]{Faculty of Fundamental Problems of Technology, Wroclaw University of Science and Technology,\newline Wybrze\.{z}e S. Wyspia\'{n}skiego 27, 50-372 Wroc\l{}aw, Poland}

\affil[+]{these authors contributed equally to this work}

\affil[*]{krzysztof.graczyk@uwr.edu.pl}

\begin{abstract}
The statistical properties of the density map (DM) approach to counting microbiological objects on images are studied in detail. The DM is given by U$^2$-Net. Two statistical methods for deep neural networks are utilized: the bootstrap and the Monte Carlo (MC) dropout. The detailed analysis of the uncertainties for the DM predictions leads to a deeper understanding of the DM model's deficiencies. Based on our investigation, we propose a self-normalization module in the network. The improved network model, called \textit{Self-Normalized Density Map} (SNDM), can correct its output density map by itself to accurately predict the total number of objects in the image. The SNDM architecture outperforms the original model. Moreover,  both statistical frameworks -- bootstrap and MC dropout -- have consistent statistical results for SNDM, which were not observed in the original model. The SNDM efficiency is comparable with the detector-base models, such as Faster and Cascade  R-CNN detectors.      
\end{abstract}

\begin{document}

\flushbottom
\maketitle
\thispagestyle{empty}

\section{Introduction}

An inseparable part of every statistical data analysis is the discussion of the uncertainties that characterize the data and the model's predictions. The first type of uncertainty is essential when formulating a model, building the likelihood, etc., while the other type reflects the model's predictive abilities. Indeed, providing the model's predictions with uncertainties allows assessing how confident the model is. A detailed error analysis of the model's predictions enables one to investigate the quality of the model, particularly verifying whether the model over- or under-fits the data. The latter aspect is crucial if one makes predictions outside the data domain.

Deep learning (DL) methods~\cite{LeCun2015} proved their excellence in many domains of life~\cite{1986Natur.323..533R}. Indeed, DL systems are utilized in image recognition and classification~\cite{NIPS2012_c399862d} as well as speech recognition~\cite{amodei2015deep} problems, or even in the game GO~\cite{Silver2016}. DL systems help optimize complex computational systems~\cite{Graczyk2020}, extracting non-trivial information hidden in big data~\cite{Najafabadi2015}. DL systems are used in high-risk applications, such as autonomous driving~\cite{Pek2020} or medical image analysis~\cite{NAIR2020101557,Moen2019,Wang2020}.
In high-risk applications, the knowledge of how certain the system's predictions are is decidedly of great importance. 

There is no doubt that proper estimation of uncertainties of a DL system's prediction's is crucial to control the validity of its model and applicability. Therefore, many groups have recently been studying this problem -- for a recent review, see Gawlikowski \textit{et al.}~\cite{gawlikowski2021survey}. As many model parameters define a DL system, optimizing it requires immense computational power. Hence the successful statistical approach to the DL system, one which takes into account various sources of uncertainties, should be simple to implement and, at the same time, should not decrease the efficiency of the resulting DL model. 

The methods for estimating the uncertainties for (shallow) neural networks (NN) were developed before the origin of deep learning. One of the directions was to use the Bayesian statistics that offer a consistent approach to error analysis for NNs~\cite{Bishop_book}.    Unfortunately, the Bayesian tools are not readily applicable to DL systems. For instance, the Bayesian technique developed by MacKay~\cite{MacKay_thesis} utilizes the Laplace approximation, which is simple in implementation and fast in execution for the shallow NN, but very difficult to implement for DL systems. Indeed, in MacKay's approach to predicting the model uncertainties, the hessian matrix must be inverted, which is impossible to perform in the case of a DL system due to too many model parameters. Other Bayesian methods utilize the Monte Carlo (MC) chain algorithms~\cite{Neal:1995}. Again, using this type of approach for the DL is difficult due to the massive number of model parameters. Nevertheless, the Bayesian-inspired, as well as the non-Bayesian statistical techniques for DL, have been developed recently~\cite{caldeira2020deeply,ovadia2019trust}. They enable DL users to perform error analysis. Many proposed approaches are based on either the MC dropout or the bootstrap technique. The MC dropout~\cite{gal2015dropout} is an example of the Bayesian-inspired approach. In contrast, the bootstrap technique is an excellent example of the non-Bayesian method~\cite{efron1979}, which statisticians have successfully exploited for years. Presently, it is widely adapted for shallow and deep NN~\cite{osb2016deep}.

This paper aims to consider one of the practical (microbiological) applications of DL systems -- a problem of counting microorganisms on images of Petri dishes. Our goal is to perform the error analysis, which will allow us to understand the validity of the counting system and let us propose the significant model's improvements. Indeed, our main achievement is a proposal of the self-normalization mechanism, implemented as additional modules in the network architecture. A counting system with self-normalization modules works more efficiently, accurately, and in the broader domain than the vanilla model.  

There are many different approaches to object counting on images. An obvious way to face this type of problem is to build a regression model \cite{chattopadhyay2017counting, hoekendijk2021counting}. The biggest advantage of this method is that it is enough to label each image with the number of objects in it. If a dataset includes more detailed annotation with a bounding box for every object in an image, it is feasible to leverage detectors for object counting~\cite{agar, majchrowska2021deep,pawlowski2021generation}. Eventually, an autoencoder can be used to estimate the density map (DM) based on a given image, which can be later integrated to obtain the number of objects~\cite{dm1,dm2,dm3,Jiang:20,Xie_doi:10.1080/21681163.2016.1149104}. It requires objects to be labeled with the coordinates of their center. The latter two methods performed better on the dataset we worked on in our experiments. The main goal of this study is to perform a comprehensive statistical analysis for a deep learning solution. Thus, we decided to focus on the density map approach because the architecture of autoencoders is usually less complex than recent neural networks for object detection, and the methods introduced in this paper are more transparent.

Counting microbial colonies on Petri dishes is an essential step in a microbiological laboratory to evaluate the cleanliness of the samples. Traditionally, this task is done manually or semi-automatically using traditional computer vision methods~\cite{semiautomat,otsu}. However, recent studies prove that a DL-based methodology accelerates the process~\cite{agar,pawlowski2021generation,beznik2020deep,Jiang:20,juhas,dnn}.
We shall utilize the DM method to predict the number of microbial colonies on a Petri dish. The DM system is defined by U$^2$-Net DL model\cite{U2net_Qin2020}. To test the DM models, we utilize the AGAR (Annotated Germs for Automated Recognition) dataset~\cite{agar}, which includes several thousand pictures of Petri dishes with five species of microbial   colonies grown.

\begin{figure}[!t]
\centering
\includegraphics[width=0.99\linewidth]{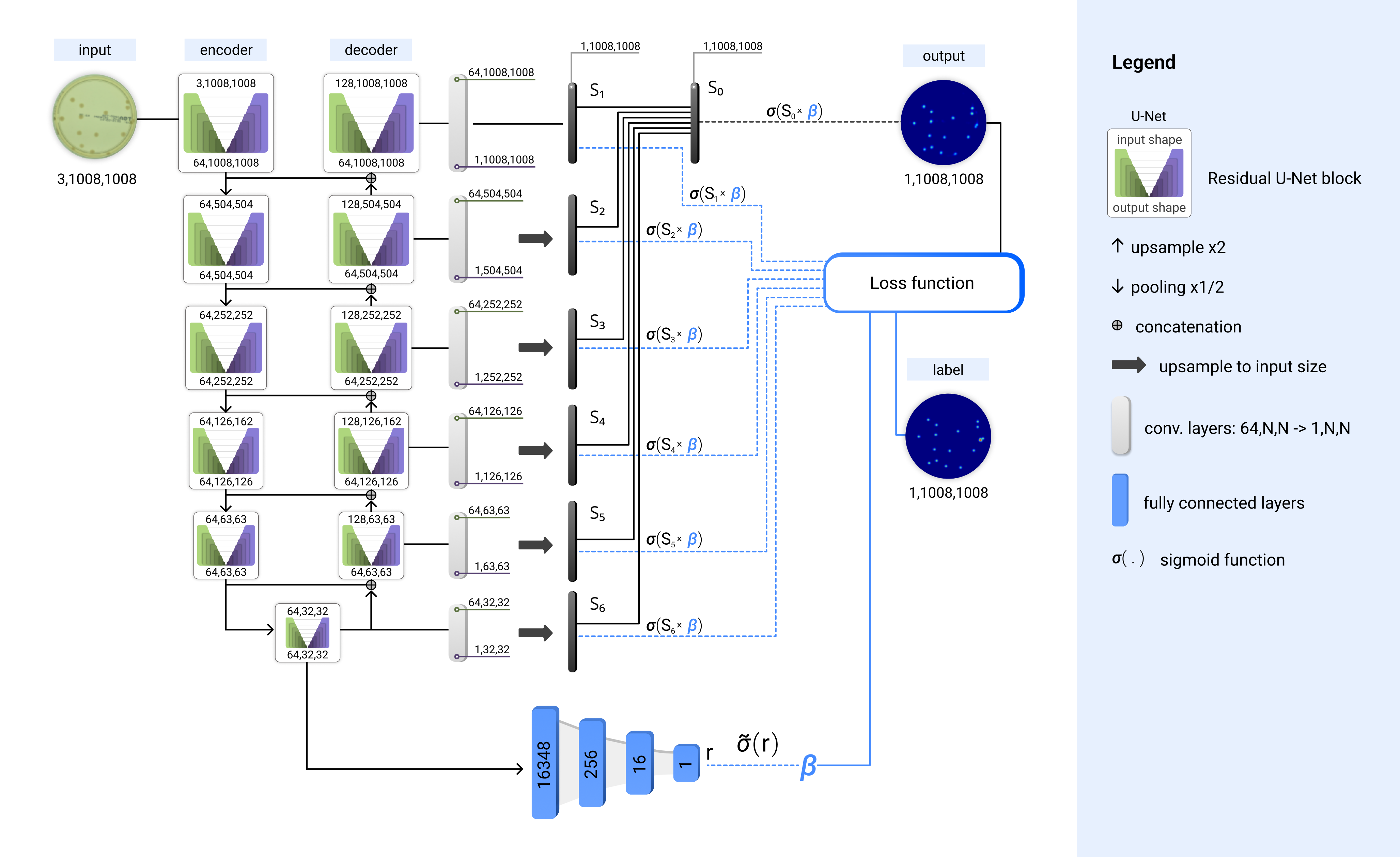}
\caption{SNDM network: U$^2$-Net extended by the normalization bypass-layers (blue) that output $\beta$ normalization parameter. Note that $\beta$ parameter multiplies last layers of the U$^2$-Net that creates the output density map.}
\label{fig:scheme}
\end{figure}

The vanilla U$^2$-Net network accurately predicts the number of microbes up to about 60 colonies but underestimates the counts above. It also overestimates its prediction for some samples with just a few colonies. We adapt two known statistical approaches to perform an error analysis of the U$^2$-Net predictions. Namely, we implement the bootstrap and MC dropout frameworks, respectively. 
The analysis of the model's uncertainties indicates that the main problem with the performance of U$^2$-Net is not the localization of microorganisms itself, but rather the normalization of density maps for a large range of possible outcomes (from 0 to 100 colonies on a Petri dish). Thus, we propose a modification of the U$^2$-Net, called \textit{Self-Normalized Density Map} (SNDM) architecture, which includes extra normalization layers significantly improving the system's performance in the previously failing domain. Moreover, the SNDM works as efficiently as detector-base approaches, such as Faster\cite{faster} and Cascade\cite{cascade} R-CNN.

The paper is organized as follows: in Section~\ref{Sec:DM} the DM and U$^2$-Net architecture is introduced and the AGAR dataset is described; Section~\ref{Sec:Error} reviews two statistical approaches for neural networks designed to estimate the uncertainties, namely, bootstrap and MC dropout; in Section~\ref{Sec:Experiments} we present our numerical results. The SNDM is introduced, and its performance is studied in detail. Conclusions are drawn in the last section.

\section{Density map and microbiological data}
\label{Sec:DM}

\subsection{U$^2$-Net as Density Map}

The DM, denoted by $\mathcal{M}$, transforms the given input image, $\mathbf{x}$, into a density map target
\begin{equation}
\mathbf{y} =   \mathcal{M} (\mathbf{x}).
\end{equation}
Note that $\mathcal{M}$ is parametrized by weights and hyperparameters, which we intentionally omit in notation because they are not relevant to our discussion. In our approach, the total number of objects in the image is obtained by summing over the entries (the pixels) of the density map $\textbf{y}$. 

The convolutional neural network (CNN) has been designed to face problems such as image recognition, image classification~\cite{Lecun98,nair2010rectified}, and segmentation. Hence the CNN-based models are also proposed to transform the image into a density map. Among several types of CNNs, architecture U-Net-type~\cite{ronneberger2015unet} is the most successful in reproducing the density map of the objects~\cite{Xie_doi:10.1080/21681163.2016.1149104}. In the present paper, we discuss the results of the analysis in which the U$^2$-Net~\cite{U2net_Qin2020} is utilized.  

The U$^2$-Net has a form of a two-level nested U-shaped structure as presented in Fig.~\ref{fig:scheme}. The top-level is built of eleven U-Net blocks, connected by pooling (encoder branch) or upsampling layers (decoder branch), with additional concatenation connections. On the bottom level, each U-Net block itself also has an encoder-decoder structure, within additional residual connections in the U-Net block. 

The analysis is performed on the AGAR dataset, including images of Petri dishes with bounding box annotations for each colony. To generate density maps, we assume that the center of objects lies in the center of the bounding box, and corresponding coordinates are added to an empty map in the form of single pixels. Next, the map is blurred with a Gaussian filter, which leads to a smooth density map normalized so that the sum over all pixels gives the total number of colonies. In that way we create the annotation map for each input image. The autoencoder is trained to transform a real image into a density map.
Note that each density map is normalized so that the sum over all pixels gives the total number of objects. Hence the network predicting the density map for an input picture also provides information about the total number of objects.

\begin{figure}[!b]
\begin{center}
\includegraphics[width=0.8\linewidth]{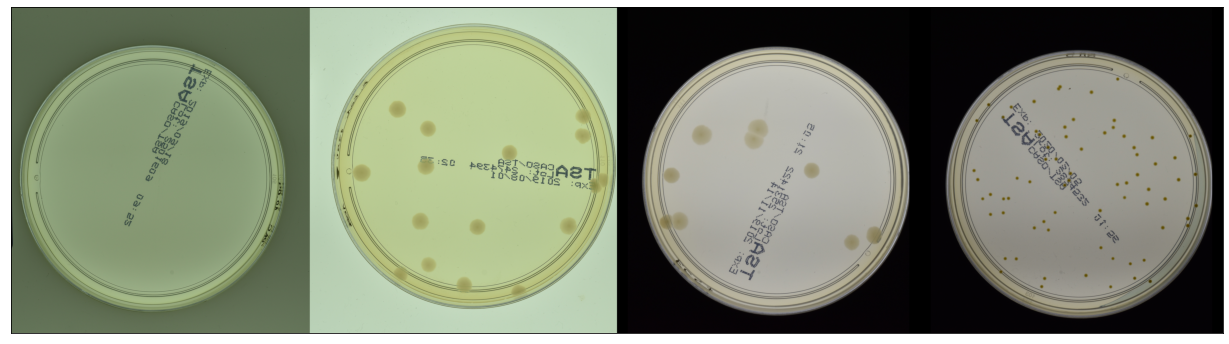}\end{center}
\caption{Exemplary images of Petri dish from the \textit{higher-resolution} subset of the AGAR dataset that was used during the numerical experiments. From the left: empty dish with no colonies, dish containing 16 colonies of \textit{B.~subtilis}, dish containing~$9$ colonies of \textit{E.~coli}, and dish containing~$72$ colonies of \textit{S.~aureus}. Samples differ in numbers of colonies, types of microbial species cultured, and image acquisition setups -- detailed information can be found in Supplementary Material for Majchrowska~\textit{et al.}\cite{agar}.}
\label{fig:samples}
\end{figure}

\begin{figure}[t!]
\begin{center}
\includegraphics[width=0.85\linewidth]{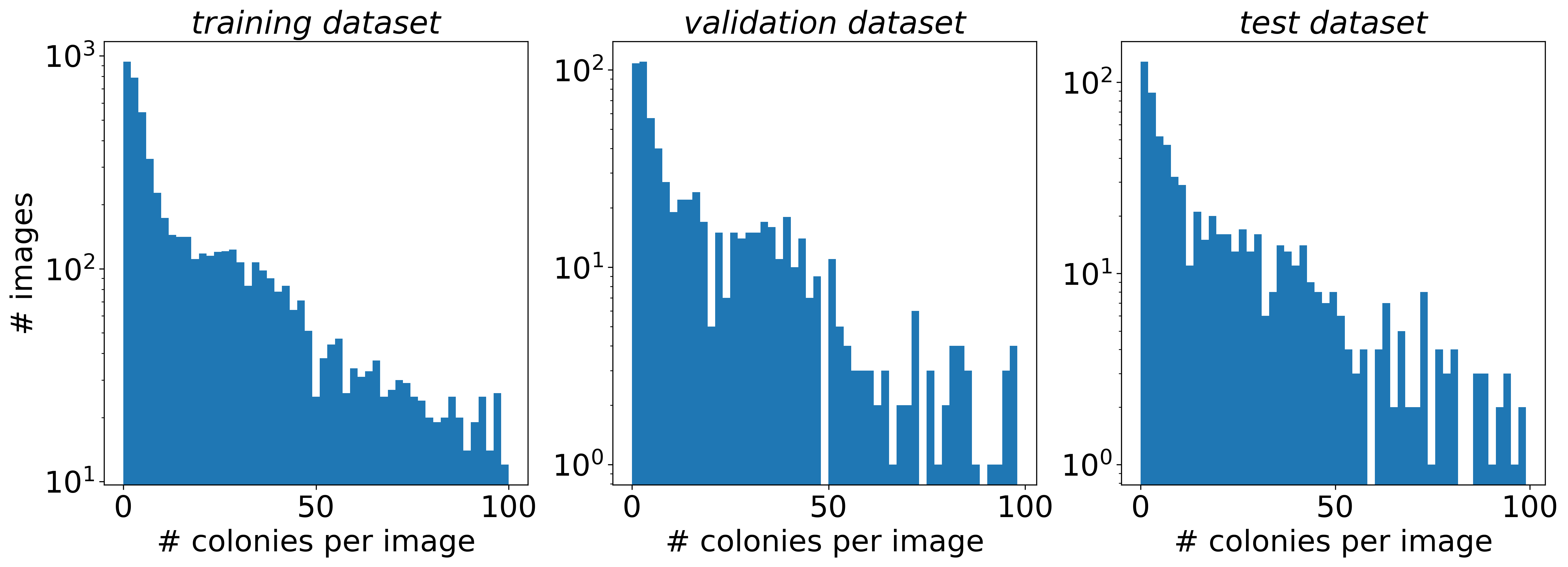}
\end{center}
\caption{Histogram of the number of annotated microbial colonies per image in the training (left), validation (center), and test (right) subsets.}
\label{fig:histogram}
\end{figure}

\subsection{AGAR dataset}
\label{Sec:Data}
As mentioned before, we utilize the AGAR dataset\cite{agar} to test the properties of the U$^2$-Net in the MC dropout and bootstrap approaches. It is a huge microbiological database including $18~000$ high-resolution images of Petri dishes with cultured colonies of $5$ standard microbes taken under diverse lighting conditions. Professional microbiologists manually annotated microbial colonies by precisely labeling each colony using a \textit{bounding box}. AGAR is diverse data collected to automate microbial colony counting. The data covers edge cases, including highly crowded plates with overlapping colonies, cases with extensive and tiny colonies, and cases where colonies are hardly visible, e.g., when located close to the dish edge. Typical plates captured in the various setups (three different illumination conditions) with microbial colonies of different shapes and sizes are shown in Fig.~\ref{fig:samples}. 

Note that AGAR mainly contains images with fewer than $50$ colonies, about $85\%$ of samples (see Fig.~\ref{fig:histogram}), or even empty dishes, which are essential for testing the models' tendency to generate false-positive counts.

\section{Uncertainties in DM}
\label{Sec:Error}

Some methods of estimation of uncertainties in the predictions of the DM, for the U-Net architecture, have recently been discussed by Eaton-Rosen~\textit{et al.}~\cite{10.1007/978-3-030-32251-9_39}. Our paper focuses on two approaches: bootstrap and MC dropout adapted for U$^2$-Net. We briefly review both of them in the following two subsections.

\subsection{Bootstrap method}

A bootstrap method for error estimation is one of the oldest and the most popular methods of statistics~\cite{efron1979}. An extensive review can be found in the work of Hastie \& Tibshirani \& Friedman~\cite{hastie01statisticallearning}, whereas Tibshirani~\cite{Tibshirani1996ACO} discusses the application of the bootstrap technique in the neural network analyses. Two types of bootstrap approaches are distinguished, namely, pairs sampling and residual sampling algorithms. We adapt the first option.

The main idea of the bootstrap is to train the same model on some number of data subsets sampled (with replacement) from the original set. Then, as a result, one obtains the ensemble of models. In the inference mode, one makes predictions of all models in the ensemble. The bootstrap model prediction is then given by the mean over the model predictions in the ensemble. Additionally, the variance over the models' outcomes gives $1\sigma$ uncertainty. 

The procedure of taking the average over the models is robust to overfitting. In our applications, we consider $B=20$ bootstrap training subsets. The same network architecture is trained for each subset, and the best model is selected using the validation subset. If the noise of the data is random and not correlated (from sample to sample), then the total error obtained for the ensemble of models is $B$ times smaller than the error of one model. For the detailed explanation, see Bishop~\cite{Bishop07}.
Note that the bootstrap approach is not a Bayesian approach, but if it is combined with early stopping, it is interpreted as approximate inference~\cite{pmlr-v51-duvenaud16}.

The known strengths of the bootstrap approach are the following:
\begin{itemize}
	\item It is simple in implementation and relatively fast in execution, i.e., it is usually enough to run the analysis for $B=10$ to $20$. 
	\item The obtained model is given by the ensemble of networks of the same architecture but with a different configuration of network parameters. 
	\item The predictions of the bootstrap model are similar to the Gaussian process (GP) in data range~\cite{osb2016deep}. Indeed, the GP model is defined by some density distribution used to sample the model. The best model prediction is given by the average over the outcomes of the model samples.
\end{itemize}
The mean overall models give the response of the bootstrap model. Therefore the bootstrap model should not overfit the data. Indeed, a single fit obtained for a particular subset of the data might be overtrained, but the mean over the models is not. Hence the bootstrap model should be characterized by a good generalization ability. 

\textit{The main steps of the bootstrap approach:}
%\begin{table}
\begin{enumerate}[label=(\roman*)]
%\begin{enumerate}
\item consider the training data set $\mathcal{D}_{train}$ with the total number of samples equals $N$;

\item obtain $B$ subsets $C_i \subset \mathcal{D}_{train}$ with an equal number of images. $C_i$ is obtained by sampling from the data $\mathcal{D}$ (with replacement) $b$ images. Note that in our experiments $B=20$, moreover $b \approx 0.63 \,N$ as it is recommended in the literature\cite{hastie01statisticallearning};

\item consider network $\mathcal{M}$ and  train it on $C_i$ data subset, as the result one gets the model $\mathcal{M}_i^B \equiv \mathcal{M}(C_i)$;   
\item to make the predictions for given input $\mathbf{x}$ compute the outcome of each network $\mathcal{M}_i^B$ and take the mean, namely:
\begin{equation}
\overline{\mathcal{M}}(\mathbf{x})    = \frac{1}{B}\sum_{i=1}^B \mathcal{M}_i(\mathbf{x}),
\end{equation}

\item to estimate the uncertainty for the prediction of the network calculate the variance:
\begin{equation}
    \Delta^2 \mathcal{M}(\mathbf{x}) = \displaystyle \frac{1}{B} \sum_{i=1}^B\left(\overline{\mathcal{M}}(\mathbf{x}) - \mathcal{M}_i(\mathbf{x})\right)^2,  
\end{equation}
where $\sqrt{\Delta^2 \mathcal{M}(\mathbf{x})}$ is interpreted as $1\sigma$ uncertainty. 
\end{enumerate}
%\end{table}

\subsection{MC Dropout}

In 1995, Neal proved that a class of neural networks with a single hidden layer and an infinite number of units converge to the Gaussian process~\cite{Neal:1995}. Moreover, Rasmussen and Williams~\cite{Rasmussen_Willimas_book} studied Gaussian process methods in supervised learning. In both cases, the Bayesian statistics stand a background for all derivations. It is believed that the GP approach allows accessing the model prediction uncertainties with reasonable accuracy.  

Motivated by the Bayesian statistics as well as the success of GP techniques in the estimation of model uncertainties, Gal and Ghahramani studied the dropout technique~\cite{gal2015dropout,gal2015dropout-appendix}. They showed that a deep neural network with a dropout layer after each weight layer can be understood as the Gaussian process. 

The main idea of the approach is to keep dropout layers active in the training and inference modes. It means that, for a given input $\mathbf{x}$, every prediction is computed with a slightly different configuration of active units. Therefore, to obtain the MC dropout prediction for a given neural network, one has to run it $N$-times. Then, the mean over the predictions gives the best value while the variance estimates the uncertainty of the predictions. 
\begin{figure}[t]
\centering
\includegraphics[width=0.75\linewidth]{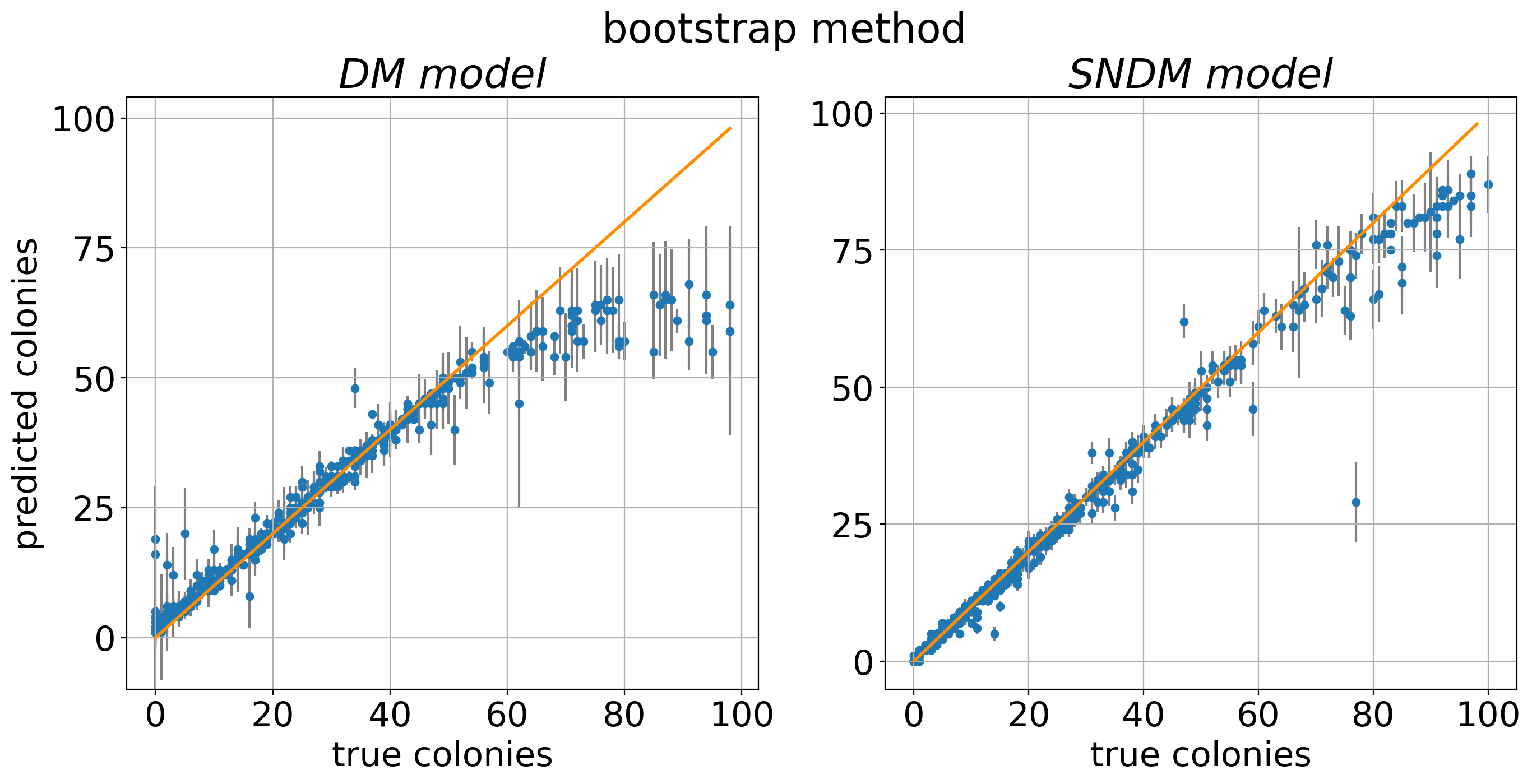}
\includegraphics[width=0.75\linewidth]{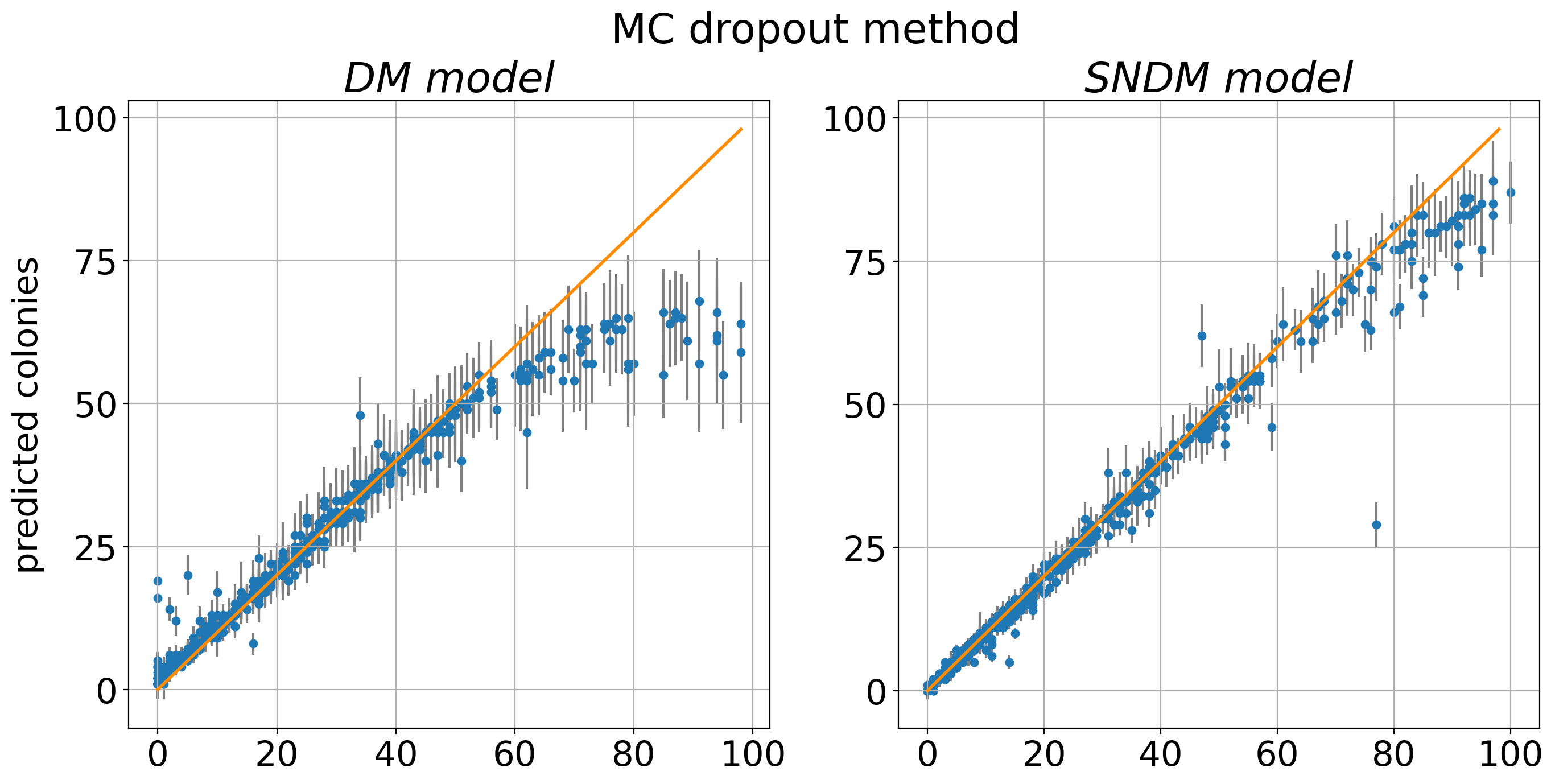}
\caption{Dependence of the predicted number of colonies on ground truth for the test data set within the DM model (right), and self-normalized DM model (left). Note that for ideal detection every blue dot representing a single Petri dish image should lay on the $y=x$ orange line. Uncertainties, denoted by black bars, in the bootstrap and MC dropout approaches are given in top and bottom rows, respectively.}
\label{fig:counting}
\end{figure}

\begin{figure}[!t]
\centering
\includegraphics[width=0.9\textwidth]{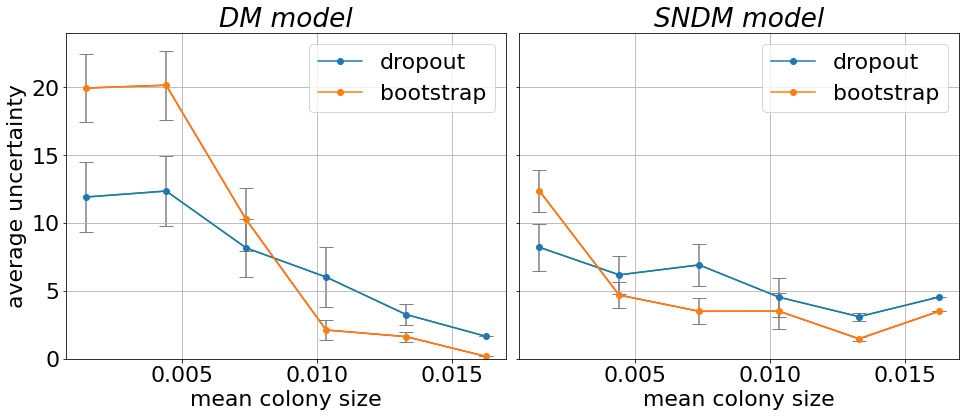}
\caption{Dependence of the counting uncertainty on mean colony size calculated for the test subset within both models. Uncertainty values are averaged on intervals of mean colonies size.}
\label{fig:err_colsize_plain}
\end{figure}
\begin{figure}[!b]
\centering
\includegraphics[width=0.9\linewidth]{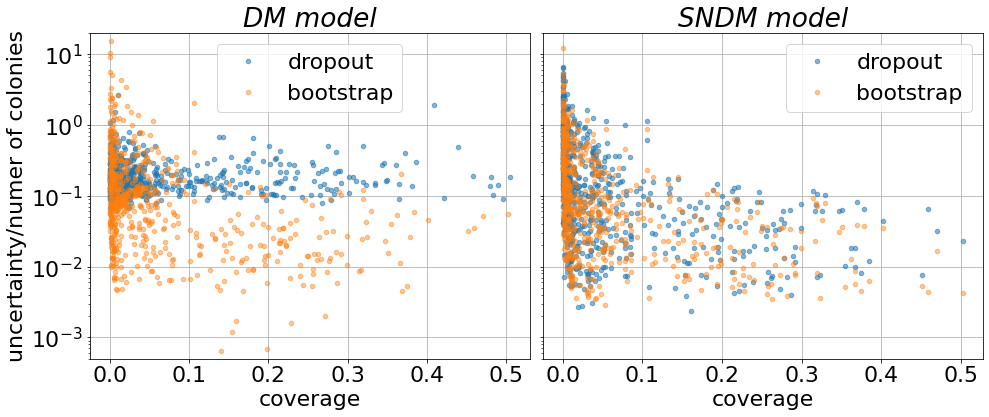}
\caption{Dependence of the counting uncertainty per single colony on coverage factor calculated for the test subset using both models.}
\label{fig:err_percolony_plain}
\end{figure}

\textit{The main steps of the MC dropout:}
\begin{enumerate}[label=(\roman*)]
%\begin{itemize}
    \item Consider a network $\mathcal{M}_{dropout}$;
    
    \item Train the $\mathcal{M}_{dropout}$ on the data set $\mathcal{D}_{train}$. The training is run with regularization -- we apply Adam algorithm with weight decay $\mu=0.0005$. Note that 
    during the training the dropout layers are active.
    
    \item To make a prediction for given input vector $\mathbf{x}$, run the network $\mathcal{M}_{dropout}$, $r=20$ times, keeping dropout layers active. Then the dropout network prediction is given by the 
    \begin{equation}
\overline{\mathcal{M}}_{dropout}(\mathbf{x})    = \frac{1}{r}\sum_{i=1}^r \mathcal{M}_{dropout}^i(\mathbf{x}),
\end{equation}
where $\mathcal{M}_{dropout}^i(\mathbf{x})$ denotes the $i$-th run prediction of the network. The $1\sigma$ uncertainty is given by the square root of the variance:
\begin{equation}
    \Delta^2 \mathcal{M}_{dropout}(\mathbf{x}) = \displaystyle \frac{1}{r} \sum_{i=1}^r\left(\overline{\mathcal{M}}_{dropout}(\mathbf{x}) - \mathcal{M}_{dropout}^i(\mathbf{x})\right)^2.  
\end{equation}
%\end{itemize}
\end{enumerate}

\begin{figure}[!b]
\centering
\includegraphics[width=0.8\linewidth]{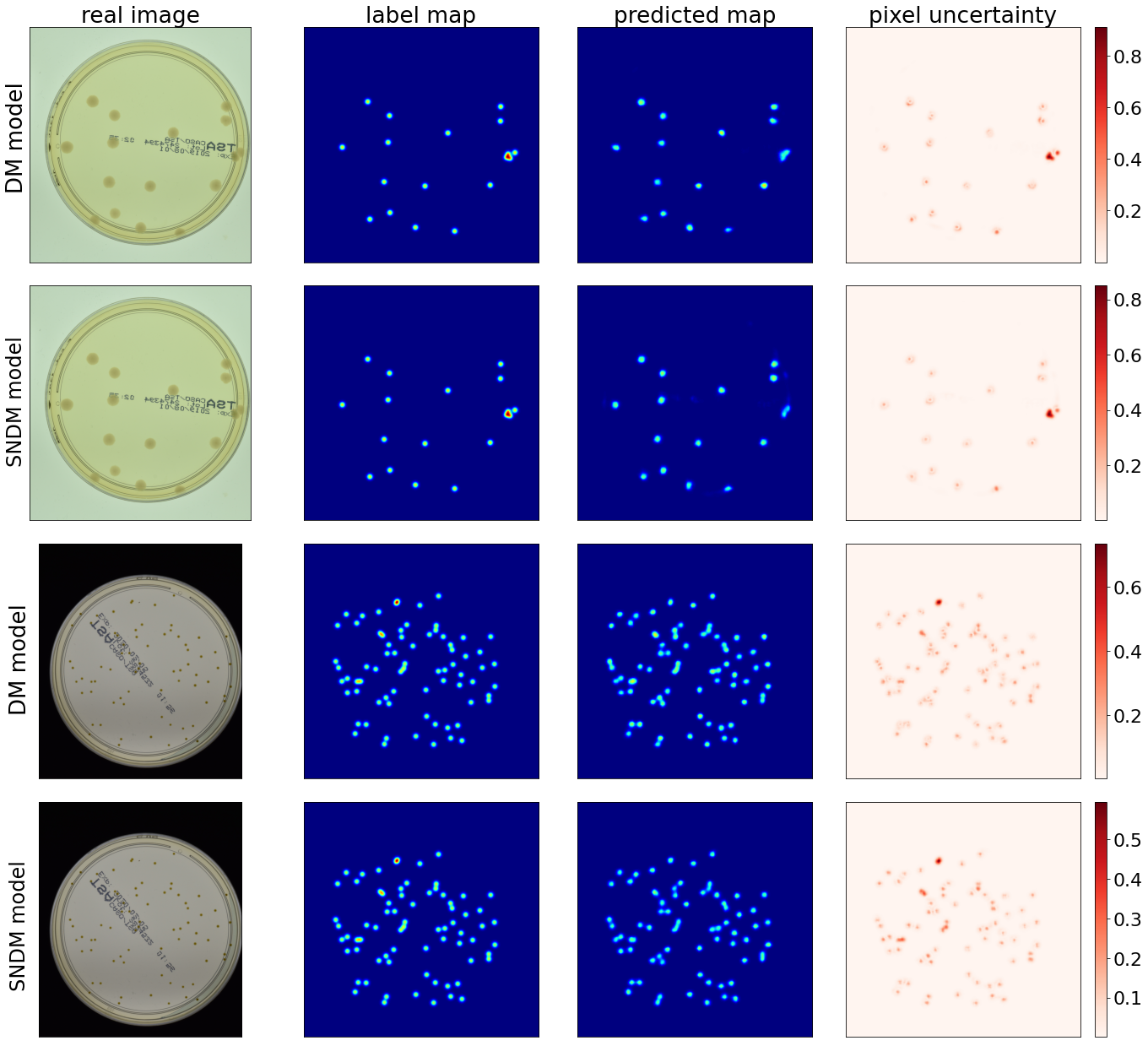}
\caption{Examples of density maps for two exemplary input images from the test subset (with $16$ and $72$ labelled colonies), together with their uncertainties per pixel: the DM model (rows 1st and 3rd), and the SNDM model (rows 2nd and 4th). In the first case DM model predicts $17\pm0.99$, while SNDM predicts $16\pm0.44$ colonies. In the second case DM predicts $63\pm3.32$, and the SNDM predicts $71\pm1.57$ colonies. Uncertainty maps calculated using MC dropout method.}
\label{fig:dmaps}
\end{figure}
\begin{figure}[!b]
\centering
\includegraphics[width=0.95\linewidth]{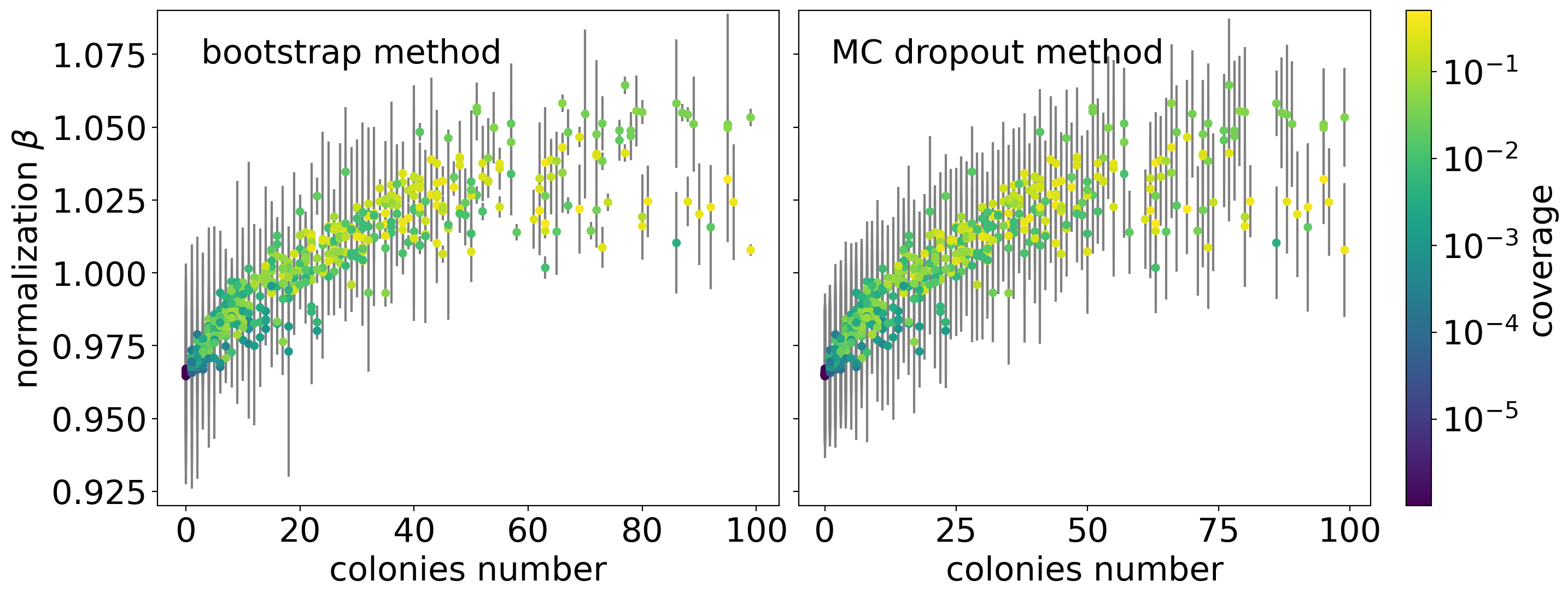}
\caption{Dependence of the normalization parameter $\beta$ on the number of colonies. Uncertainties of the $\beta$ parameter calculated using bootstrap method (left) and MC dropout (right). Results for the test subset.}
\label{fig:beta}
\end{figure}

\section{Numerical results}
\label{Sec:Experiments}
This section examines the properties of the U$^2$-Net architecture as a density map predictor. The results of these studies allow us to propose the self-normalization mechanism in the U$^2$-Net model. We justify the relevance of this approach in the following sections.

To study the U$^2$-Net properties, we consider the \textit{higher-resolution} subset of the AGAR dataset, represented by RGB images of Petri dishes with a resolution of about $4000\times4000$~pixels. We limit our discussion to images containing no more than $100$ colonies. Because, from the microbiological point of view, the accuracy of automatic counting is less relevant for more populated samples, i.e. determining whether the sample is contaminated or not is the most important.

The analyzed data is divided into three subsets: training ($5656$ images), validation ($706$ images), and test ($706$ images) data subset. As introduced earlier, we consider two types of statistical approaches for estimating uncertainties in predictions of deep neural networks: the bootstrap and the MC dropout. In both methods, we exploit the same network architecture shown in Fig.~\ref{fig:scheme}. Moreover, the same training scheme is adapted for both models -- the Adam optimization algorithm with \textit{learning rate} of $0.0001$, \textit{weight decay} equals $0.0005$, and the size of the minibatch of two. The training takes at least $100$ epochs.

We ran the computations on NVIDIA TITAN RTX GPU with $24$~GB of memory. The training time was about $40$~h for MC dropout. The bootstrap experiments took longer, and the training lasted about four days, performed in parallel on five GPUs.

\subsection{Performance of vanilla U$^2$-Net as a density map predictor}

To access the quality of both discussed models, we compare the predictions of the network versus the ground truth computed on the test data set; see the left column in Fig.~\ref{fig:counting}. Each predicted number of colonies is given with $1\sigma$ uncertainty. Note that both statistical models work well for the samples with the number of colonies smaller than $60$. However, for images with the number of colonies lower than $10$, the DM count is overestimated. Predictions with tiny uncertainties characterize the MC dropout for these cases.
Consequently, the model disagrees with the data within the $1\sigma$ confidence level. In contrast, the bootstrap model predictions have larger uncertainties. Hence, the model agrees with the ground truth in the $1\sigma$ level. From that perspective, the bootstrap approach works better than the MC dropout. 
\begin{table}[]
\centering
\begin{tabular}{l|cc|cc|}
\cline{2-5}
& \multicolumn{2}{c|}{bootstrap} & \multicolumn{2}{c|}{MC dropout}                       \\ \hline
\multicolumn{1}{|l|}{dishes with:}      & \multicolumn{1}{c|}{DM}      & SNDM   & \multicolumn{1}{l|}{DM}   & \multicolumn{1}{l|}{SNDM} \\ \hline
\multicolumn{1}{|l|}{$\le50$ colonies} & \multicolumn{1}{c|}{1.26}    & 0.80   & \multicolumn{1}{c|}{2.38} & 1.31                      \\ \hline
\multicolumn{1}{|l|}{$>50$ colonies}   & \multicolumn{1}{c|}{5.67}    & 3.94   & \multicolumn{1}{c|}{7.74} & 5.01                      \\ \hline
\multicolumn{1}{|l|}{overall}          & \multicolumn{1}{c|}{1.76}    & 1.19   & \multicolumn{1}{c|}{2.99} & 1.78                      \\ \hline
\end{tabular}
\caption{Average uncertainties of predictions for the test dataset within the DM and the SNDM models in the bootstrap and MC dropout approaches (as presented in Fig.~\ref{fig:counting}). Uncertainties averaged in a group of less (more) crowded dishes with $\le50$ ($>50$) colonies and overall.}
\label{tab:avg_unc}
\end{table}

The detailed examination of our results shows that the uncertainty of the model's predictions depends on the mean size of colonies for a given dish. It is illustrated in Fig.~\ref{fig:err_colsize_plain} (left panel). Interestingly, for the images with smaller colonies, the uncertainties for the bootstrap approach are larger than for the dropout model. However, when the size of the colonies increases, the effect is the opposite.

We noted that the uncertainties of the model's prediction (normalized to the number of colonies) weakly depend on the dish coverage (the fraction of the dish covered by colonies), as shown in Fig.~\ref{fig:err_percolony_plain} (left panel). It is interesting to note that the MC dropout uncertainties are systematically larger than for the bootstrap approach, see Table \ref{tab:avg_unc}, where the average (over the test dataset) uncertainties are given.  

In Fig.~\ref{fig:dmaps}, we showed four panels. Each contains an input picture, a map of annotations, network prediction, and a map of uncertainties computed for each output pixel. We show the results for two dishes that contain $16$ and $72$ bacterial colonies, respectively. The DM prediction agrees with the ground truth in the first case (first panel). In contrast to the second case (third panel), the DM underestimates the count. However, in the latter case, the DM correctly reproduces spots representing the objects in the image, but the most significant uncertainties are consistently observed in the spot positions. Therefore, we conclude that the DM reconstructs the map of spots (objects) correctly but has a problem finding the correct normalization for the images with a larger number of colonies. Therefore, we propose a modification of the network architecture so that the network can correct by itself the normalization of the output. 

\subsection{Self-Normalized DM (SNDM)}

To propose the normalization module in the network, we must explain how the loss function is constructed in the U$^2$-Net model.

Let us denote by $E_{\mathrm{U}^2}$ the loss function U$^2$-Net. It is a function of the data and model parameters. From the statistical perspective, it refers to the logarithm of the likelihood function. In practice, $E_{\mathrm{U}^2}$ has a quite complicated structure\cite{U2net_Qin2020}:
\begin{equation}
\label{Eq:loss_withh_normalization_penalty}
E_{\mathrm{U}^2} = \sum_\mathrm{images} \sum_{M=0}^6 \mathcal{L}( P_M,P^G)
\end{equation}
where $P_M$ and $P^G$ denote predicted density map by $M$-th U-Net block and the ground truth, respectively.

The $\mathcal{L}_M$ is the standard binary cross entropy function:
\begin{equation}
    \mathcal{L}(P,P_G) = - \sum_{i,j}^{H,W} \left[  {P_G}_{i,j} \ln  P_{i,j} +  
    (1 - {P_G}_{i,j}) \ln  (1- P_{i,j}) \right],
\end{equation}
where $i,j$ denotes the pixel indexes in the figure of the width $W$ and height $H$.

The U$^2$-Net generates six output maps, denoted by $S_{1,\dots,6}$, from the encoder ($S_6$) and five from the decoder branch ($S_{1,\dots,5}$), respectively. Outputs from smaller blocks ($S_{2,\dots,6}$) are next upsampled to fit the size of the output map. Note that $S_1$ already has the proper $1008\times1008$~size. After upsampling (we still denote them by $S_i$), each of them gives the density map and $P_i = \sigma(S_i)$, where $\sigma$ is the sigmoid activation function. All maps $P_{1..6}$ contributes to the loss function. Moreover, $S_{1..6}$ maps are fused with a concatenation operation followed by a $1\!\times\!1$ convolution layer (giving $S_0$). Then $P_0=\sigma(S_0)$ is the final density map output of the U$^2$-Net. The $P_0$ contributes also to the loss function.

Our idea is to allow the network to correct the normalization of output pixels in $S_i$ maps. Since every output pixel ranges from $0$ to $1$, one can not simply re-scale the $P_i$-outputs by some normalization number. Therefore, we propose the rescaling of pixels normalization in $S_i$ map. We use the property that sigmoid is a linear function on neighborhood, namely, $\sigma(x) \approx = \frac{1}{2} + \frac{x}{2} +\dots$. Hence, the rescaling of the sigmoid argument changes the total normalization of the sigmoid almost linearly in this range. In practice, we rescale each $S_i$ output by a factor $\beta$, so
\begin{equation}
    P_i = \sigma(S_i) \to \sigma(\beta S_i). 
\end{equation}

\begin{figure}[!t]
\begin{center}
\includegraphics[width=0.6\linewidth]{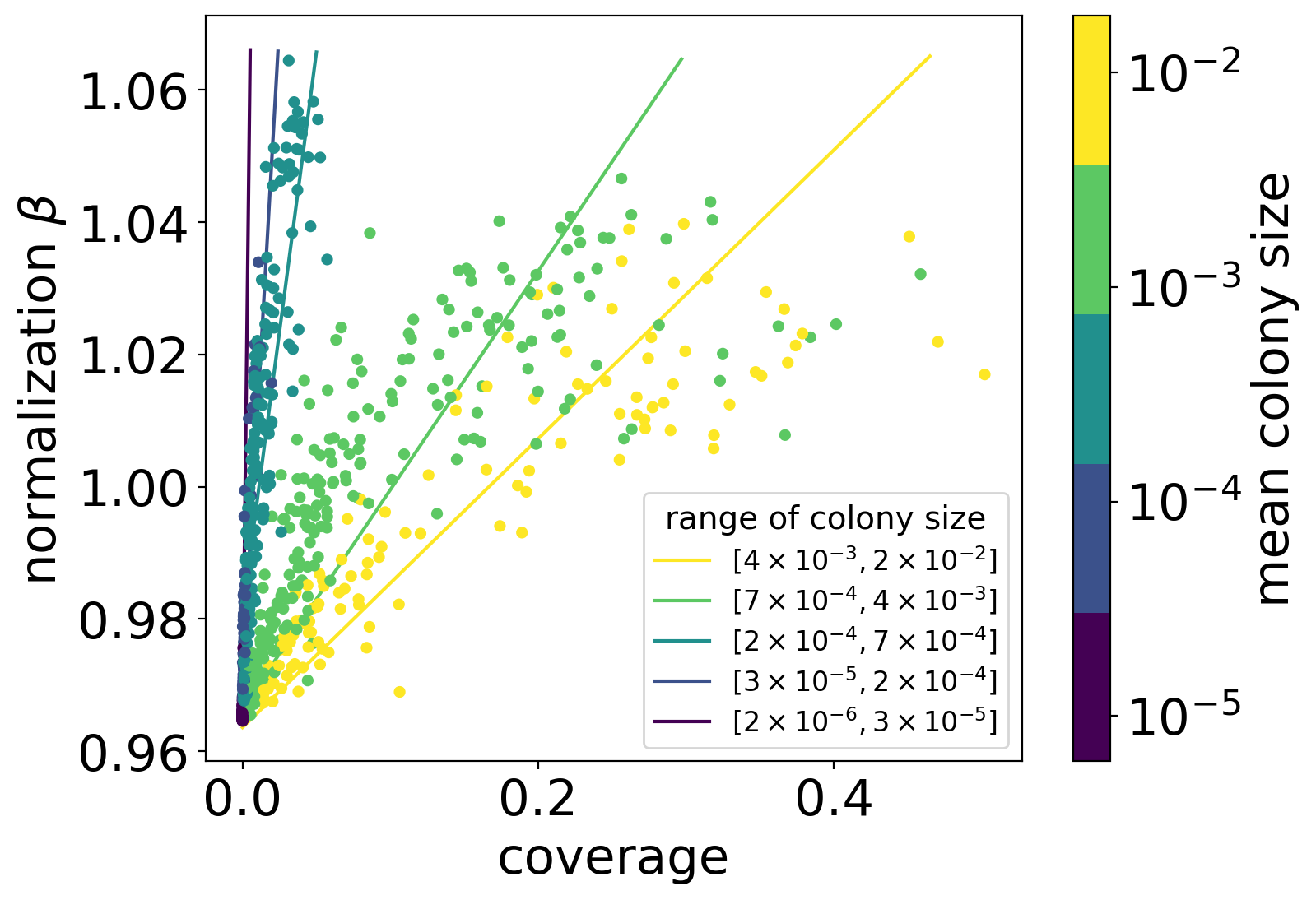}
\end{center}
\caption{Dependence of $\beta$ parameter on dish coverage, together with mean colony size indicated by the color of points. The points for different ranges of colony size (see the legend) cluster in subsets that show approximately linear dependence between $\beta$ and coverage. Additionally, a linear function is fitted in each range, with a slope that decreases with colony size. Results obtained within MC dropout for the test subset.}
\label{fig:beta_lin}
\end{figure}

The normalization value $\beta$ might depart from one, and it can be a function of the distribution of colonies, the total coverage, etc. Therefore, we assume that $\beta$ depends on the input image and is an additional network outcome. To limit the number of parameters that define $\beta$ we connect it with the smallest $U$-Net block of U$^2$-Net, the encoder part, see Fig.~\ref{fig:scheme}. In the rest of the paper, the U$^2$-Net architecture with our modification will be called a \textit{self-normalized density map} (SNDM).

In principle, the system should work without the need for corrections, $\beta=1$, hence we expect that $\beta$ should take values around one. To keep the control $\beta$-dependence, we add to the loss term a penalty term: 
\begin{equation}
\label{Eq:new_lose}
    %\mathcal{L}(P,P_G) \to \mathcal{L}(P,P_G)
    E_{\mathrm{U}^2} \to E_{\mathrm{U}^2}+\sum_{images}\frac{1}{2}\left(1 - \beta \right)^2.
\end{equation}

In general $\beta $ can take any positive number, but, in practical applications, we assume that $\beta = \tilde \sigma (\dots) = \lambda \sigma(\dots)$ to speed up the process of training. It guarantees that normalization ranges from $0$ to $1.5$. In our experiments, we tested a sigmoid with $\lambda=2$ and $\tilde\sigma=ReLU$. However, the most optimal results have been obtained for the sigmoid activation function with $\lambda=1.5$.

In Fig.~\ref{fig:counting} (right column of panel), we plot the predictions of SNDM versus ground truth. We see that modifying U$^2$-Net architecture by introducing the self-normalization module significantly improves the network's predictions. Indeed, the SNDM works well for images containing less than $60$ objects and samples with the number of objects larger than $60$. Interestingly, in the improved model, the bootstrap and MC dropout predictions are in agreement. The estimated uncertainties for both approaches became similar, see  Figs.~\ref{fig:err_colsize_plain} and \ref{fig:err_percolony_plain} (right panel), as well as Table~\ref{tab:avg_unc}. Moreover, the bootstrap uncertainties for SNDM are reduced with respect to the original model.

\begin{figure}[!b]
\centering
\includegraphics[width=0.95\linewidth]{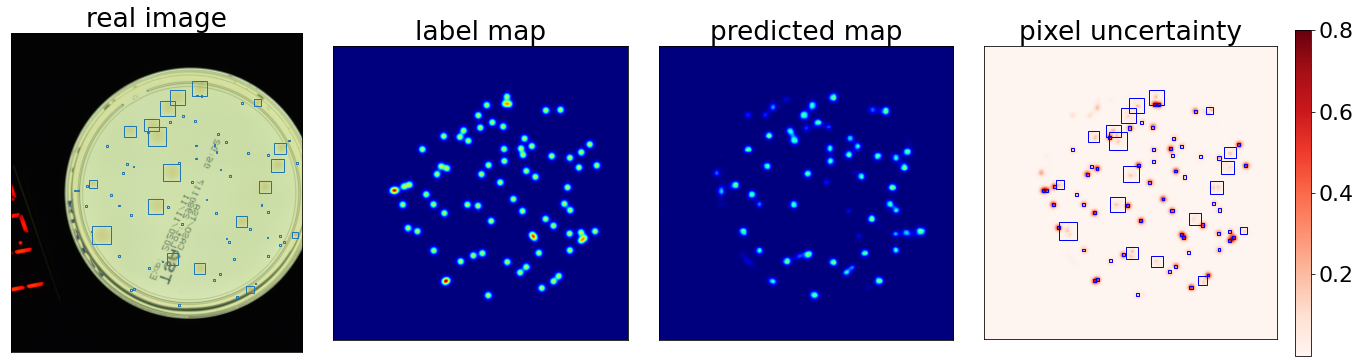}
\caption{Density maps of one of the outliers from Fig.~\ref{fig:counting} predicted with the SNDM. The model predicts $29\pm7.30$ colonies, while the sample contains  $77$ colonies of two types: $19$ \textit{P. aeruginosa} (big ones), and 58 \textit{C. albicans} (small ones). The blue boxes represent all the colonies' ground truth position and size. Uncertainties calculated using the bootstrap method.}
\label{fig:outlier}
\end{figure}

The normalization parameter $\beta$ depends on the features of the input image, such as the number of colonies. It is illustrated in Fig.~\ref{fig:beta}. For both statistical approaches, $\beta$ varies from $0.96$ to $1.06$ with the tendency to take lower values for less crowded images. On the other hand, for the images containing more than $50$ objects, $\beta$ varies less. 
It is also worth mentioning that $\beta$ takes a wide range of values for a given number of colonies, which varies with the number of colonies, which translates to coverage and other features like mean colony size.
It is visible in Fig.~\ref{fig:beta_lin} where we plot the dependence of $\beta$ on the dish coverage. When we group the samples by the mean colony size (indicated by different colors of points in Fig.~\ref{fig:beta_lin}), one can observe that $\beta$ almost linearly depends on coverage. Moreover, the slope of the fitted lines increases gradually with the mean size of colonies.

Fig.~\ref{fig:dmaps}, shows density maps together with pixel uncertainties for two input pictures with $16$ and $72$ objects, respectively. The predictions for the SNDM are shown in the second and the fourth panel, respectively. We see that SNDM outperforms the original U$^2$-Net. Indeed, the SNDM accurately predicts the number of colonies for both images.

Eventually, we note that, overlapping objects is one of the difficulties the counter system must face. In Fig.~\ref{fig:dmaps}, we see an example of colonies that overlap. In this case, the amplitude of density maps is naturally enlarged. But the interesting thing is that also uncertainties have higher values there -- see the pixel uncertainty maps in the last column of Fig.~\ref{fig:dmaps}. Therefore, we may conclude that these areas have a more substantial impact on counting uncertainties. However, if one compares DM and SNDM pixel uncertainty in the overlapping regions, then in the SNDM case, uncertainties are slightly reduced compared to DM.

The detailed inspection of the results presented in Fig.~\ref{fig:counting} shows that there are input figures for which the SNDM network overshoots. In Fig.~\ref{fig:outlier} we plot the predicted map with uncertainties for such a case. We see that the network's main problem is the presence of colonies of different sizes on the same dish. It complicates the inference of the normalization parameter, which also depends on the size of the colonies, see in Fig.~\ref{fig:beta_lin}. The blue boxes in Fig.~\ref{fig:outlier} represent ground truth labels showing the precise position and size of the colonies. It should be also noted that the pixel uncertainties for small colonies in the small box area are much larger than for big colonies located in bigger boxes. 

\begin{table}[]
\centering
\begin{tabular}{c|c|c|}
\cline{2-3}
& \multicolumn{2}{c|}{\cellcolor[HTML]{EFEFEF}detection metrics} \\ \hline
\rowcolor[HTML]{EFEFEF} 
\multicolumn{1}{|c|}{\cellcolor[HTML]{EFEFEF}model} & MAE & sMAPE \\ \hline
\rowcolor[HTML]{FFFFFF} 
\multicolumn{1}{|c|}{DM} & 5.74 & 12.58\% \\ \hline
\rowcolor[HTML]{FFFFFF} 
\multicolumn{1}{|c|}{\begin{tabular}[c]{@{}c@{}}self-normalized \\ DM\end{tabular}} & 3.65 & 4.05\% \\ \hline
\rowcolor[HTML]{FFFFFF} 
\multicolumn{1}{|c|}{Faster R-CNN} & 4.75 & 5.32\% \\ \hline
\rowcolor[HTML]{FFFFFF} 
\multicolumn{1}{|c|}{\begin{tabular}[c]{@{}c@{}}Cascade \\ R-CNN \end{tabular}} & 4.31 & 4.86 \% \\ \hline
\end{tabular}
\caption{Counting metrics, Mean Absolute Error (MAE) (Eq.~\ref{Eq:MAE}) and symmetric Mean Absolute Percentage Error (sMAPE) (Eq.~\ref{Eq:sMAE}), for the DM and SNDM model (in MC dropout) in comparison with other counting methods that use popular R-CNN object detectors: Faster\cite{faster} and Cascade\cite{cascade}, which are discussed by Majchrowska \textit{et al.}\cite{agar}. }
\label{tab:meth_comp}
\end{table}

We close the discussion of the numerical result by comparing the SNDM model predictions with the other popular counting systems.  
In one of our previous papers\cite{agar,pawlowski2021generation} we discussed some of them. For the comparison, we choose two of the most successful approaches (in our experiments) to count AGAR microbiological colonies, namely, Faster R-CNN detector\cite{faster} and Cascade R-CNN detector\cite{cascade}. Note that, for both models, the algorithm first performs the detection of interesting objects and then counts them.

To quantitatively compare various models predictions, we consider a standard Mean Absolute Error:
\begin{equation}
\label{Eq:MAE}
\mathrm{MAE} =\frac{1}{N}\sum_{i=1}^N|n_i - \Tilde{n}_i|
,
\end{equation}
and less common in usage, symmetric Mean Absolute Percentage Error:
\begin{equation}
\label{Eq:sMAE} \mathrm{sMAPE} = \frac{100\%}{N}\sum_{i=1}^N\frac{|n_i - \Tilde{n}_i|}{|n_i + \Tilde{n}_i|},
\end{equation}
where $N$ is a number of all samples, $n_i$ 
is true count of microbe colonies present on $i$-th image, and $\Tilde{n}_i$ is predicted number of colonies. Note that the sMAPE shows the relative error with respect to the overall number of colonies.

Both metrics are obtained for the \textit{higher-resolution} subset of the AGAR dataset and the four models: Faster and Cascade detectors, as well as DM and SNDM in the MC dropout approach. The metrics are given in Table~\ref{tab:meth_comp}.  We see that counting by detection gives better results than the standard DM model. But the SNDM gives comparable or even slightly better counting results than the detectors. Further discussion of other deep learning detectors, e.g., well-known YOLO, can be found in our previous paper\cite{majchrowska2021deep}. Note that preparing annotations for the density map models are less expensive than for detector models. Moreover, the training of the detector systems is more resources-demanding and time-consuming than the density map models.   Therefore, for the data for which  the density map approaches work with the same or better efficiency as the detectors, the density map systems as SNDM are recommended.

\section{Summary}

We have shown that the detailed analysis of the uncertainties in the deep neural network model leads to a deeper understanding of its limitations. It allows us to assess how uncertain the predictions are. More profound knowledge of the model allowed us to modify the DM model by introducing the Self-Normalization mechanism. The SNDM significantly improves the DM predictions. Indeed, before the modification, the model underestimated the number of colonies for highly populated dishes.
In contrast, for some images with a small number of colonies, the DM model overestimated the counts. The SNDM network proposed by us can correctly count in both mentioned domains. Its performance is comparable with the other counting systems, such as Faster and Cascade  R-CNN detectors.
Eventually, we note that the bootstrap and the MC dropout predictions and the estimate of uncertainties in the SNDM are in statistical agreement, which was not observed in the original model.

\section*{Availability of Data and Materials}
The AGAR dataset used during the current study is available from \url{https://agar.neurosys.com/} on reasonable request.

\section*{Acknowledgements}
Project “Development of a new method for detection and identifying bacterial colonies using artificial neural networks and machine learning algorithms” is co-financed from European Union funds under the European Regional Development Funds as part of the Smart Growth Operational Program. Project implemented as part of the National Centre for Research and Development: Fast Track (grant no. POIR.01.01.01-00-0040/18).\\
K.M.G. has been partly supported by the program ''Excellence initiative - research university'' for the years 2020-2026 for the University of  Wroc\l{}aw, as well as by MOZART grant founded by Wroclaw Academic Center.

\bibliographystyle{alpha}
\bibliography{sample}

\begin{thebibliography}{10}
\urlstyle{rm}
\expandafter\ifx\csname url\endcsname\relax
  \def\url#1{\texttt{#1}}\fi
\expandafter\ifx\csname urlprefix\endcsname\relax\def\urlprefix{URL }\fi
\expandafter\ifx\csname doiprefix\endcsname\relax\def\doiprefix{DOI: }\fi
\providecommand{\bibinfo}[2]{#2}
\providecommand{\eprint}[2][]{\url{#2}}

\bibitem{LeCun2015}
\bibinfo{author}{LeCun, Y.}, \bibinfo{author}{Bengio, Y.} \&
  \bibinfo{author}{Hinton, G.}
\newblock \bibinfo{journal}{\bibinfo{title}{Deep learning}}.
\newblock {\emph{\JournalTitle{Nature}}} \textbf{\bibinfo{volume}{521}},
  \bibinfo{pages}{436 EP --} (\bibinfo{year}{2015}).

\bibitem{1986Natur.323..533R}
\bibinfo{author}{Rumelhart, D.~E.}, \bibinfo{author}{Hinton, G.~E.} \&
  \bibinfo{author}{Williams, R.~J.}
\newblock \bibinfo{journal}{\bibinfo{title}{{Learning representations by
  back-propagating errors}}}.
\newblock {\emph{\JournalTitle{Nature}}} \textbf{\bibinfo{volume}{323}},
  \bibinfo{pages}{533--536}, \doiprefix\url{10.1038/323533a0}
  (\bibinfo{year}{1986}).

\bibitem{NIPS2012_c399862d}
\bibinfo{author}{Krizhevsky, A.}, \bibinfo{author}{Sutskever, I.} \&
  \bibinfo{author}{Hinton, G.~E.}
\newblock \bibinfo{title}{Imagenet classification with deep convolutional
  neural networks}.
\newblock In \bibinfo{editor}{Pereira, F.}, \bibinfo{editor}{Burges, C. J.~C.},
  \bibinfo{editor}{Bottou, L.} \& \bibinfo{editor}{Weinberger, K.~Q.} (eds.)
  \emph{\bibinfo{booktitle}{Advances in Neural Information Processing
  Systems}}, vol.~\bibinfo{volume}{25} (\bibinfo{publisher}{Curran Associates,
  Inc.}, \bibinfo{year}{2012}).

\bibitem{amodei2015deep}
\bibinfo{author}{Amodei, D.} \emph{et~al.}
\newblock \bibinfo{title}{Deep speech 2: End-to-end speech recognition in
  english and mandarin} (\bibinfo{year}{2015}).
\newblock \eprint{1512.02595}.

\bibitem{Silver2016}
\bibinfo{author}{Silver, D.} \emph{et~al.}
\newblock \bibinfo{journal}{\bibinfo{title}{Mastering the game of go with deep
  neural networks and tree search}}.
\newblock {\emph{\JournalTitle{Nature}}} \textbf{\bibinfo{volume}{529}},
  \bibinfo{pages}{484--489}, \doiprefix\url{10.1038/nature16961}
  (\bibinfo{year}{2016}).

\bibitem{Graczyk2020}
\bibinfo{author}{Graczyk, K.~M.} \& \bibinfo{author}{Matyka, M.}
\newblock \bibinfo{journal}{\bibinfo{title}{Predicting porosity, permeability,
  and tortuosity of porous media from images by deep learning}}.
\newblock {\emph{\JournalTitle{Scientific Reports}}}
  \textbf{\bibinfo{volume}{10}}, \bibinfo{pages}{21488},
  \doiprefix\url{10.1038/s41598-020-78415-x} (\bibinfo{year}{2020}).

\bibitem{Najafabadi2015}
\bibinfo{author}{Najafabadi, M.~M.} \emph{et~al.}
\newblock \bibinfo{journal}{\bibinfo{title}{Deep learning applications and
  challenges in big data analytics}}.
\newblock {\emph{\JournalTitle{Journal of Big Data}}}
  \textbf{\bibinfo{volume}{2}}, \bibinfo{pages}{1},
  \doiprefix\url{10.1186/s40537-014-0007-7} (\bibinfo{year}{2015}).

\bibitem{Pek2020}
\bibinfo{author}{Pek, C.}, \bibinfo{author}{Manzinger, S.},
  \bibinfo{author}{Koschi, M.} \& \bibinfo{author}{Althoff, M.}
\newblock \bibinfo{journal}{\bibinfo{title}{Using online verification to
  prevent autonomous vehicles from causing accidents}}.
\newblock {\emph{\JournalTitle{Nature Machine Intelligence}}}
  \textbf{\bibinfo{volume}{2}}, \bibinfo{pages}{518--528},
  \doiprefix\url{10.1038/s42256-020-0225-y} (\bibinfo{year}{2020}).

\bibitem{NAIR2020101557}
\bibinfo{author}{Nair, T.}, \bibinfo{author}{Precup, D.},
  \bibinfo{author}{Arnold, D.~L.} \& \bibinfo{author}{Arbel, T.}
\newblock \bibinfo{journal}{\bibinfo{title}{Exploring uncertainty measures in
  deep networks for multiple sclerosis lesion detection and segmentation}}.
\newblock {\emph{\JournalTitle{Medical Image Analysis}}}
  \textbf{\bibinfo{volume}{59}}, \bibinfo{pages}{101557},
  \doiprefix\url{https://doi.org/10.1016/j.media.2019.101557}
  (\bibinfo{year}{2020}).

\bibitem{Moen2019}
\bibinfo{author}{Moen, E.} \emph{et~al.}
\newblock \bibinfo{journal}{\bibinfo{title}{Deep learning for cellular image
  analysis}}.
\newblock {\emph{\JournalTitle{Nature Methods}}} \textbf{\bibinfo{volume}{16}},
  \bibinfo{pages}{1233--1246}, \doiprefix\url{10.1038/s41592-019-0403-1}
  (\bibinfo{year}{2019}).

\bibitem{Wang2020}
\bibinfo{author}{Wang, G.}, \bibinfo{author}{Ye, J.~C.} \&
  \bibinfo{author}{De~Man, B.}
\newblock \bibinfo{journal}{\bibinfo{title}{Deep learning for tomographic image
  reconstruction}}.
\newblock {\emph{\JournalTitle{Nature Machine Intelligence}}}
  \textbf{\bibinfo{volume}{2}}, \bibinfo{pages}{737--748},
  \doiprefix\url{10.1038/s42256-020-00273-z} (\bibinfo{year}{2020}).

\bibitem{gawlikowski2021survey}
\bibinfo{author}{Gawlikowski, J.} \emph{et~al.}
\newblock \bibinfo{title}{A survey of uncertainty in deep neural networks}
  (\bibinfo{year}{2021}).
\newblock \eprint{2107.03342}.

\bibitem{Bishop_book}
\bibinfo{author}{{C. M., Bishop}}.
\newblock \emph{\bibinfo{title}{{Neural Networks for Pattern Recognition}}}
  (\bibinfo{publisher}{Oxford University Press}, \bibinfo{year}{1995}).

\bibitem{MacKay_thesis}
\bibinfo{author}{MacKay, D.}
\newblock \emph{\bibinfo{title}{Bayesian Methods for Adaptive Models}}.
\newblock Ph.D. thesis, \bibinfo{school}{California Institute of Technology}
  (\bibinfo{year}{1991}).

\bibitem{Neal:1995}
\bibinfo{author}{Neal, R.~M.}
\newblock \emph{\bibinfo{title}{Bayesian learning for neural networks}}.
\newblock Ph.D. thesis, \bibinfo{school}{Graduate Department of Computer
  Science in University of Toronto} (\bibinfo{year}{1995}).

\bibitem{caldeira2020deeply}
\bibinfo{author}{Caldeira, J.} \& \bibinfo{author}{Nord, B.}
\newblock \bibinfo{title}{Deeply uncertain: Comparing methods of uncertainty
  quantification in deep learning algorithms} (\bibinfo{year}{2020}).
\newblock \eprint{2004.10710}.

\bibitem{ovadia2019trust}
\bibinfo{author}{Ovadia, Y.} \emph{et~al.}
\newblock \bibinfo{title}{Can you trust your model's uncertainty? evaluating
  predictive uncertainty under dataset shift} (\bibinfo{year}{2019}).
\newblock \eprint{1906.02530}.

\bibitem{gal2015dropout}
\bibinfo{author}{Gal, Y.} \& \bibinfo{author}{Ghahramani, Z.}
\newblock \bibinfo{title}{Dropout as a bayesian approximation: Representing
  model uncertainty in deep learning} (\bibinfo{year}{2015}).
\newblock \eprint{1506.02142}.

\bibitem{efron1979}
\bibinfo{author}{Efron, B.}
\newblock \bibinfo{journal}{\bibinfo{title}{Bootstrap methods: Another look at
  the jackknife}}.
\newblock {\emph{\JournalTitle{Ann. Statist.}}} \textbf{\bibinfo{volume}{7}},
  \bibinfo{pages}{1--26}, \doiprefix\url{10.1214/aos/1176344552}
  (\bibinfo{year}{1979}).

\bibitem{osb2016deep}
\bibinfo{author}{Osband, I.}, \bibinfo{author}{Blundell, C.},
  \bibinfo{author}{Pritzel, A.} \& \bibinfo{author}{Roy, B.~V.}
\newblock \bibinfo{title}{Deep exploration via bootstrapped dqn}
  (\bibinfo{year}{2016}).
\newblock \eprint{1602.04621}.

\bibitem{chattopadhyay2017counting}
\bibinfo{author}{Chattopadhyay, P.}, \bibinfo{author}{Vedantam, R.},
  \bibinfo{author}{Selvaraju, R.~R.}, \bibinfo{author}{Batra, D.} \&
  \bibinfo{author}{Parikh, D.}
\newblock \bibinfo{title}{Counting everyday objects in everyday scenes}.
\newblock In \emph{\bibinfo{booktitle}{Proceedings of the IEEE conference on
  computer vision and pattern recognition}}, \bibinfo{pages}{1135--1144}
  (\bibinfo{year}{2017}).

\bibitem{hoekendijk2021counting}
\bibinfo{author}{Hoekendijk, J.} \emph{et~al.}
\newblock \bibinfo{journal}{\bibinfo{title}{Counting using deep learning
  regression gives value to ecological surveys}}.
\newblock {\emph{\JournalTitle{Scientific reports}}}
  \textbf{\bibinfo{volume}{11}}, \bibinfo{pages}{1--12} (\bibinfo{year}{2021}).

\bibitem{agar}
\bibinfo{author}{Majchrowska, S.} \emph{et~al.}
\newblock \bibinfo{journal}{\bibinfo{title}{{AGAR} a microbial colony dataset
  for deep learning detection}}.
\newblock {\emph{\JournalTitle{arXiv preprint arXiv:2108.01234}}}
  (\bibinfo{year}{2021}).

\bibitem{majchrowska2021deep}
\bibinfo{author}{Majchrowska, S.} \emph{et~al.}
\newblock \bibinfo{journal}{\bibinfo{title}{Deep neural networks approach to
  microbial colony detection--a comparative analysis}}.
\newblock {\emph{\JournalTitle{arXiv preprint arXiv:2108.10103}}}
  (\bibinfo{year}{2021}).

\bibitem{pawlowski2021generation}
\bibinfo{author}{Paw{\l}owski, J.}, \bibinfo{author}{Majchrowska, S.} \&
  \bibinfo{author}{Golan, T.}
\newblock \bibinfo{journal}{\bibinfo{title}{Generation of microbial colonies
  dataset with deep learning style transfer}}.
\newblock {\emph{\JournalTitle{Scientific Reports}}}
  \textbf{\bibinfo{volume}{12}}, \bibinfo{pages}{5212},
  \doiprefix\url{10.1038/s41598-022-09264-z} (\bibinfo{year}{2022}).

\bibitem{dm1}
\bibinfo{author}{Lempitsky, V.} \& \bibinfo{author}{Zisserman, A.}
\newblock \bibinfo{journal}{\bibinfo{title}{Learning to count objects in
  images}}.
\newblock {\emph{\JournalTitle{Advances in neural information processing
  systems}}} \textbf{\bibinfo{volume}{23}}.

\bibitem{dm2}
\bibinfo{author}{Arteta, C.}, \bibinfo{author}{Lempitsky, V.},
  \bibinfo{author}{Noble, J.~A.} \& \bibinfo{author}{Zisserman, A.}
\newblock \bibinfo{title}{Interactive object counting}.
\newblock In \emph{\bibinfo{booktitle}{European conference on computer
  vision}}, \bibinfo{pages}{504--518} (\bibinfo{organization}{Springer},
  \bibinfo{year}{2014}).

\bibitem{dm3}
\bibinfo{author}{Zhang, C.}, \bibinfo{author}{Li, H.}, \bibinfo{author}{Wang,
  X.} \& \bibinfo{author}{Yang, X.}
\newblock \bibinfo{title}{Cross-scene crowd counting via deep convolutional
  neural networks}.
\newblock In \emph{\bibinfo{booktitle}{Proceedings of the IEEE conference on
  computer vision and pattern recognition}}, \bibinfo{pages}{833--841}
  (\bibinfo{year}{2015}).

\bibitem{Jiang:20}
\bibinfo{author}{Jiang, N.} \& \bibinfo{author}{Yu, F.}
\newblock \bibinfo{journal}{\bibinfo{title}{Multi-column network for cell
  counting}}.
\newblock {\emph{\JournalTitle{OSA Continuum}}} \textbf{\bibinfo{volume}{3}},
  \bibinfo{pages}{1834--1846}, \doiprefix\url{10.1364/OSAC.396603}
  (\bibinfo{year}{2020}).

\bibitem{Xie_doi:10.1080/21681163.2016.1149104}
\bibinfo{author}{Xie, W.}, \bibinfo{author}{Noble, J.~A.} \&
  \bibinfo{author}{Zisserman, A.}
\newblock \bibinfo{journal}{\bibinfo{title}{Microscopy cell counting and
  detection with fully convolutional regression networks}}.
\newblock {\emph{\JournalTitle{Computer Methods in Biomechanics and Biomedical
  Engineering: Imaging \& Visualization}}} \textbf{\bibinfo{volume}{6}},
  \bibinfo{pages}{283--292}, \doiprefix\url{10.1080/21681163.2016.1149104}
  (\bibinfo{year}{2018}).
\newblock \eprint{https://doi.org/10.1080/21681163.2016.1149104}.

\bibitem{semiautomat}
\bibinfo{author}{Selinummi, J.}, \bibinfo{author}{Sepp{\"a}l{\"a}, J.},
  \bibinfo{author}{Yli-Harja, O.} \& \bibinfo{author}{Puhakka, J.~A.}
\newblock \bibinfo{journal}{\bibinfo{title}{Software for quantification of
  labeled bacteria from digital microscope images by automated image
  analysis}}.
\newblock {\emph{\JournalTitle{Biotechniques}}} \textbf{\bibinfo{volume}{39}},
  \bibinfo{pages}{859--863} (\bibinfo{year}{2005}).

\bibitem{otsu}
\bibinfo{author}{Chen, W.-B.} \& \bibinfo{author}{Zhang, C.}
\newblock \bibinfo{journal}{\bibinfo{title}{An automated bacterial colony
  counting and classification system}}.
\newblock {\emph{\JournalTitle{Information Systems Frontiers}}}
  \textbf{\bibinfo{volume}{11}}, \bibinfo{pages}{349--368}
  (\bibinfo{year}{2009}).

\bibitem{beznik2020deep}
\bibinfo{author}{Beznik, T.}, \bibinfo{author}{Smyth, P.},
  \bibinfo{author}{de~Lannoy, G.} \& \bibinfo{author}{Lee, J.~A.}
\newblock \bibinfo{journal}{\bibinfo{title}{Deep learning to detect bacterial
  colonies for the production of vaccines}}.
\newblock {\emph{\JournalTitle{arXiv preprint arXiv:2009.00926}}}
  (\bibinfo{year}{2020}).

\bibitem{juhas}
\bibinfo{author}{Zhang, Y.}, \bibinfo{author}{Jiang, H.}, \bibinfo{author}{Ye,
  T.} \& \bibinfo{author}{Juhas, M.}
\newblock \bibinfo{journal}{\bibinfo{title}{Deep learning for imaging and
  detection of microorganisms}}.
\newblock {\emph{\JournalTitle{Trends in Microbiology}}}
  \textbf{\bibinfo{volume}{29}}, \bibinfo{pages}{569--572},
  \doiprefix\url{https://doi.org/10.1016/j.tim.2021.01.006}
  (\bibinfo{year}{2021}).

\bibitem{dnn}
\bibinfo{author}{Wang, H.} \emph{et~al.}
\newblock \bibinfo{journal}{\bibinfo{title}{Early detection and classification
  of live bacteria using time-lapse coherent imaging and deep learning}}.
\newblock {\emph{\JournalTitle{Light: Science \& Applications}}}
  \textbf{\bibinfo{volume}{9}}, \bibinfo{pages}{1--17} (\bibinfo{year}{2020}).

\bibitem{U2net_Qin2020}
\bibinfo{author}{Qin, X.} \emph{et~al.}
\newblock \bibinfo{journal}{\bibinfo{title}{U2-net: Going deeper with nested
  u-structure for salient object detection}}.
\newblock {\emph{\JournalTitle{Pattern Recognition}}}
  \textbf{\bibinfo{volume}{106}}, \bibinfo{pages}{107404},
  \doiprefix\url{10.1016/j.patcog.2020.107404} (\bibinfo{year}{2020}).

\bibitem{faster}
\bibinfo{author}{Ren, S.}, \bibinfo{author}{He, K.}, \bibinfo{author}{Girshick,
  R.} \& \bibinfo{author}{Sun, J.}
\newblock \bibinfo{journal}{\bibinfo{title}{Faster {R-CNN}: Towards real-time
  object detection with region proposal networks}}.
\newblock {\emph{\JournalTitle{Advances in neural information processing
  systems}}} \textbf{\bibinfo{volume}{28}}, \bibinfo{pages}{91--99}
  (\bibinfo{year}{2015}).

\bibitem{cascade}
\bibinfo{author}{Cai, Z.} \& \bibinfo{author}{Vasconcelos, N.}
\newblock \bibinfo{title}{Cascade {R-CNN}: Delving into high quality object
  detection}.
\newblock In \emph{\bibinfo{booktitle}{Proceedings of the IEEE conference on
  computer vision and pattern recognition}}, \bibinfo{pages}{6154--6162}
  (\bibinfo{year}{2018}).

\bibitem{Lecun98}
\bibinfo{author}{Lecun, Y.}, \bibinfo{author}{Bottou, L.},
  \bibinfo{author}{Bengio, Y.} \& \bibinfo{author}{Haffner, P.}
\newblock \bibinfo{journal}{\bibinfo{title}{Gradient-based learning applied to
  document recognition}}.
\newblock {\emph{\JournalTitle{Proceedings of the IEEE}}}
  \textbf{\bibinfo{volume}{86}}, \bibinfo{pages}{2278--2324}
  (\bibinfo{year}{1998}).

\bibitem{nair2010rectified}
\bibinfo{author}{Nair, V.} \& \bibinfo{author}{Hinton, G.~E.}
\newblock \bibinfo{title}{Rectified linear units improve restricted boltzmann
  machines}.
\newblock In \bibinfo{editor}{Fürnkranz, J.} \& \bibinfo{editor}{Joachims, T.}
  (eds.) \emph{\bibinfo{booktitle}{Proceedings of the 27th International
  Conference on Machine Learning (ICML-10)}}, \bibinfo{pages}{807--814}
  (\bibinfo{year}{2010}).

\bibitem{ronneberger2015unet}
\bibinfo{author}{Ronneberger, O.}, \bibinfo{author}{Fischer, P.} \&
  \bibinfo{author}{Brox, T.}
\newblock \bibinfo{title}{U-net: Convolutional networks for biomedical image
  segmentation} (\bibinfo{year}{2015}).
\newblock \eprint{1505.04597}.

\bibitem{10.1007/978-3-030-32251-9_39}
\bibinfo{author}{Eaton-Rosen, Z.}, \bibinfo{author}{Varsavsky, T.},
  \bibinfo{author}{Ourselin, S.} \& \bibinfo{author}{Cardoso, M.~J.}
\newblock \bibinfo{title}{As easy as 1, 2...4? uncertainty in counting tasks
  for medical imaging}.
\newblock In \bibinfo{editor}{Shen, D.} \emph{et~al.} (eds.)
  \emph{\bibinfo{booktitle}{Medical Image Computing and Computer Assisted
  Intervention -- MICCAI 2019}}, \bibinfo{pages}{356--364}
  (\bibinfo{publisher}{Springer International Publishing},
  \bibinfo{address}{Cham}, \bibinfo{year}{2019}).

\bibitem{hastie01statisticallearning}
\bibinfo{author}{Hastie, T.}, \bibinfo{author}{Tibshirani, R.} \&
  \bibinfo{author}{Friedman, J.}
\newblock \emph{\bibinfo{title}{The Elements of Statistical Learning}}.
\newblock Springer Series in Statistics (\bibinfo{publisher}{Springer New York
  Inc.}, \bibinfo{address}{New York, NY, USA}, \bibinfo{year}{2001}).

\bibitem{Tibshirani1996ACO}
\bibinfo{author}{Tibshirani, R.}
\newblock \bibinfo{journal}{\bibinfo{title}{A comparison of some error
  estimates for neural network models}}.
\newblock {\emph{\JournalTitle{Neural Computation}}}
  \textbf{\bibinfo{volume}{8}}, \bibinfo{pages}{152--163}
  (\bibinfo{year}{1996}).

\bibitem{Bishop07}
\bibinfo{author}{Bishop, C.~M.}
\newblock \emph{\bibinfo{title}{Pattern Recognition and Machine Learning}}
  (\bibinfo{publisher}{Springer}, \bibinfo{address}{New York},
  \bibinfo{year}{2007}).

\bibitem{pmlr-v51-duvenaud16}
\bibinfo{author}{Duvenaud, D.}, \bibinfo{author}{Maclaurin, D.} \&
  \bibinfo{author}{Adams, R.}
\newblock \bibinfo{title}{Early stopping as nonparametric variational
  inference}.
\newblock vol.~\bibinfo{volume}{51} of \emph{\bibinfo{series}{Proceedings of
  Machine Learning Research}}, \bibinfo{pages}{1070--1077}
  (\bibinfo{publisher}{PMLR}, \bibinfo{address}{Cadiz, Spain},
  \bibinfo{year}{2016}).

\bibitem{Rasmussen_Willimas_book}
\bibinfo{author}{Rasmussen, C.} \& \bibinfo{author}{Williams, C.}
\newblock \emph{\bibinfo{title}{Gaussian Processes for Machine Learning}}.
\newblock Adaptive Computation and Machine Learning (\bibinfo{publisher}{MIT
  Press}, \bibinfo{address}{Cambridge, MA, USA}, \bibinfo{year}{2006}).

\bibitem{gal2015dropout-appendix}
\bibinfo{author}{Gal, Y.} \& \bibinfo{author}{Ghahramani, Z.}
\newblock \bibinfo{title}{Dropout as a bayesian approximation: Appendix}
  (\bibinfo{year}{2015}).
\newblock \eprint{1506.02157}.

\end{thebibliography}
\end{document}